\documentclass[letterpaper]{article} 
\usepackage{aaai24}  
\usepackage{times}  
\usepackage{helvet}  
\usepackage{courier}  
\usepackage[hyphens]{url}  
\usepackage{graphicx} 
\usepackage{natbib}  
\usepackage{caption} 
\usepackage{algorithm}
\usepackage{algorithmic}
\usepackage{array,booktabs,siunitx}
\usepackage{multirow}
\usepackage{tabularx}

\usepackage{newfloat}
\usepackage{listings}
\DeclareCaptionStyle{ruled}{labelfont=normalfont,labelsep=colon,strut=off} 
\floatstyle{ruled}
\newfloat{listing}{tb}{lst}{}
\floatname{listing}{Listing}
\title{When To Grow? \\ A Fitting Risk-Aware Policy for Layer Growing in Deep Neural Networks}
\author{
    Haihang Wu\equalcontrib,
    Wei Wang,
    Tamasha Malepathirana,
    Damith Senanayake,\\
    Denny Oetomo,
    Saman Halgamuge
}
\affiliations{
    Department of Mechanical Engineering\\
    The University of Melbourne\\

    haihangw@student.unimelb.edu.au\,
}

\usepackage{bibentry}

\begin{document}

\maketitle

\begin{abstract}

Neural growth is the process of growing a small neural network to a large network and has been utilized to accelerate the training of deep neural networks. One crucial aspect of neural growth is determining the optimal growth timing. However, few studies investigate this systematically. Our study reveals that neural growth inherently exhibits a regularization effect, whose intensity is influenced by the chosen policy for growth timing. While this regularization effect may mitigate the overfitting risk of the model, it may lead to a notable accuracy drop when the model underfits. Yet, current approaches have not addressed this issue due to their lack of consideration of the regularization effect from neural growth. Motivated by these findings, we propose an under/over fitting risk-aware growth timing policy, which automatically adjusts the growth timing informed by the level of potential under/overfitting risks to address both risks.  Comprehensive experiments conducted using CIFAR-10/100 and ImageNet datasets show that the proposed policy achieves accuracy improvements of up to 1.3\% in models prone to underfitting while achieving similar accuracies in models suffering from overfitting compared to the existing methods.

\end{abstract}

\section{Introduction}
Deep neural networks (DNNs) have significantly advanced the state of the art in various computer vision tasks, including image classification \cite{He2016DeepRecognition, Dosovitskiy2021AnScale}, object detection \cite{Fang2021YouDetection}, and image segmentation \cite{Kirillov2023SegmentAnything}. For instance, with a model size 10 times larger than ResNet-50 \cite{He2016DeepRecognition}, the pioneering convolutional neural network ConvNext \cite{Liu2022A2020s} achieves a substantial improvement in classification accuracy on the ImageNet dataset \cite{Deng2009Imagenet:Database} compared to ResNet. However, this improvement comes at the expense of approximately 50 times  more computational burden \cite{Liu2022A2020s}. Consequently, the development of efficient training methods for DNN becomes imperative.

\begin{figure}[h]
\centering
\includegraphics[width=\linewidth]{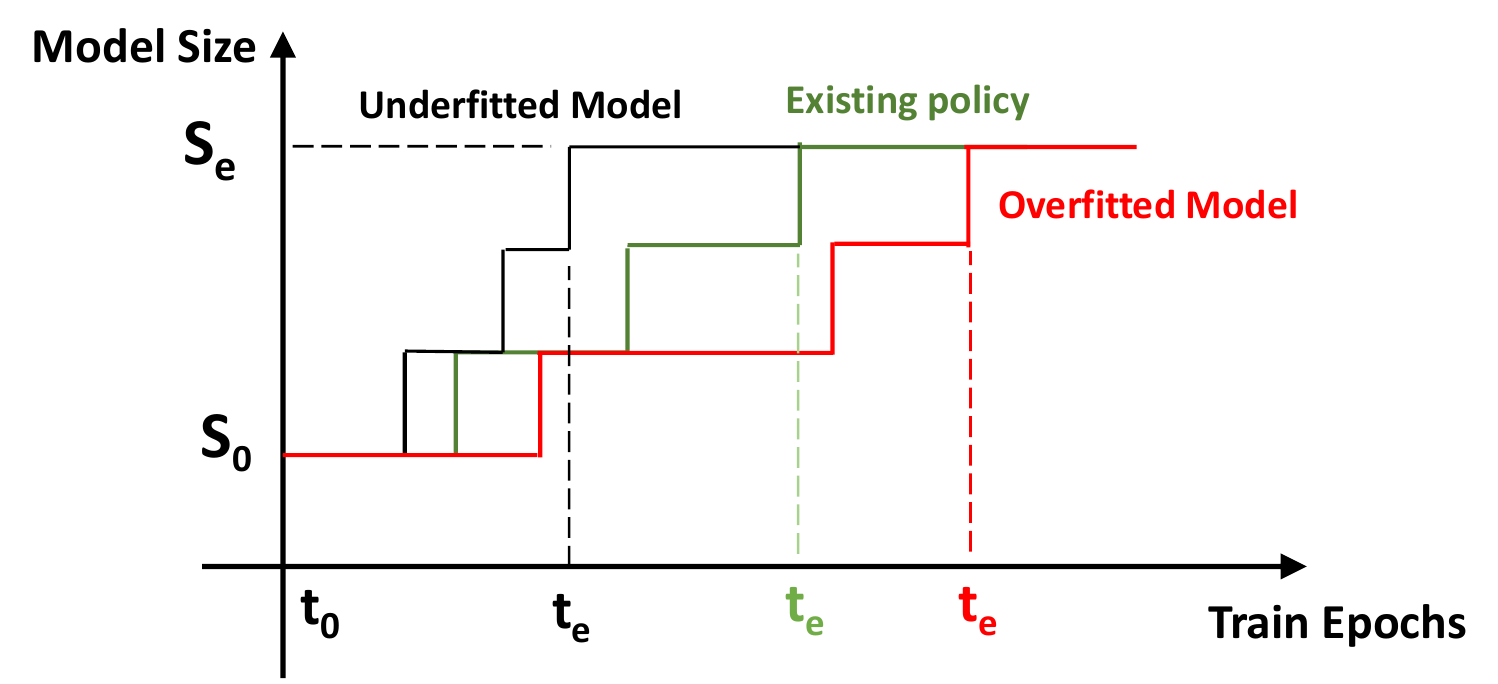}
\vspace{-0.5cm}
\caption{\textbf{When to grow policy.} To expedite the training process of the target large model with size $S_e$, neural growth grows the initial small model from size $S_0$ at time $t_0$ into  $S_e$ at time $t_e$. For the prevailing conventional strategies (illustrated by the green curve), $t_e$ is not influenced by overfitting or underfitting risks directly, and these strategies may fail to address under/over fitting risks effectively. By contrast, our fitting risk-aware policy accelerates growth (black curve) by reducing $t_e$ to mitigate underfitting risks when the target model exhibits underfitting tendencies. Conversely, it decelerates growth (red curve) by extending the $t_e$ for models displaying signs of overfitting,  addressing the overfitting risk.  }\label{intro:when to grow policy}
\vspace{-0.5cm}
\end{figure}

Existing works demonstrate that neural growth can significantly reduce training time with minimal impact on accuracy \cite{Li2022AutomatedTransformers,Chang2018Multi-levelView}. For instance, studies  \cite{Chang2018Multi-levelView,Dong2020TowardsPerspective}  show neural growth can save around 40\% training time for deep ResNet on CIFAR datasets with minor accuracy loss \cite{Krizhevsky2009LearningImages}, and a similar finding has also been observed in vision transformer growth \cite{Li2022AutomatedTransformers}. Although the study of neural growth encompasses various dimensions \cite{Maile2022WhenANNs, Maile2022StructuralPerspective}, such as initialization of new neurons/layers, when to grow (growth timing), and where to grow,  these works predominantly focus on the initialization of new neurons or layers.

Few studies \cite{Wen2020AutoGrow:Networks,Dong2020TowardsPerspective} investigate the ``when to grow" policy for neural growth. Two major policies are periodic growth which adds layers or neurons at regular intervals, and convergent growth which adds layers only when the current model has converged. In this study, we investigate the impact of neural growth on the model accuracy and then develop a ``when to grow" policy that considers this impact.

\begin{figure*}[htb]
\centering
\includegraphics[width=\linewidth]{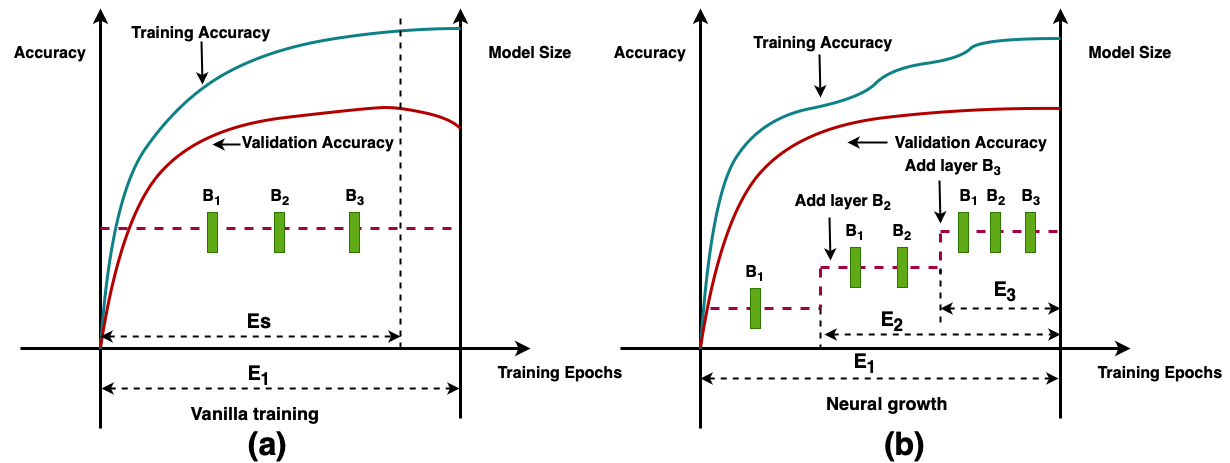}
\vspace{-0.3cm}
\caption{\textbf{Comparison of vanilla training and neural growth.} In standard training (vanilla), all layers undergo training for $E_1$ epochs. However, in the context of neural growth, only layer $B_1$ is trained for $E_1$ epochs, while the majority of layers ($B_2$ and $B_3$) are subjected to reduced training epochs ($E_2$ and $E_3$ respectively).  }\label{fig:1} 
\vspace{-0.3cm}
\end{figure*}

Our investigation reveals that neural growth inherently provides a regularization effect on the final model, whose strength is controlled by the growth timing policy. Figure ~\ref{fig:1} (b) shows that neural growth reduces the average training epochs when compared with the vanilla method without neural growth. Reduced exposure to training data prevents the model from memorizing irrelevant information (noise) in the training dataset, potentially applying regularization to the model. Moreover, the regularization strength is controlled by the average training epochs or the growth timing policy. Although this regularization effect mitigates the overfitting risk by reducing learned noise, the model might not capture sufficient signals from the training dataset, particularly when it underfits the data due to this regularization effect. Nevertheless, existing ``when to grow" policies disregard this matter due to their lack of consideration for the neural growth-induced regularization effect. Table ~\ref{when to grow policy} shows that existing growth timing policies (periodic growth and convergent growth) achieve slightly better or similar accuracy with shorter training time on the overfitting CIFAR10/100 dataset compared to the vanilla method without neural growth. However, they exhibit a significant accuracy decline on the underfitting ImageNet dataset compared to the vanilla method.

We introduce a novel \textbf{F}itting \textbf{R}isk-\textbf{A}ware \textbf{G}row policy (FRAGrow) to tackle this concern. This policy evaluates underfitting and overfitting risks at each growth phase and dynamically adjusts the growth timing to mitigate these risks. As Figure ~\ref{intro:when to grow policy} shows, our policy accelerates growth to introduce milder regularization on the final model when dealing with underfitting models. Conversely, when overfitting is detected, the policy slows down the growth rate, thereby applying stronger regularization to the final model. By doing so, our ``when to grow" policy effectively addresses the risks of overfitting and underfitting. Experimental results demonstrate that our approach achieves superior accuracy compared to existing policies, albeit with a trade-off in training time.

In summary, our contributions are:

\begin{itemize}
    \item We identify that neural growth inherently exhibits a regularization effect, the intensity of which is governed by the growth timing.
    \item Based on this observation, we propose a fitting risk-aware policy that dynamically adjusts the growth timing by evaluating fitting risks with the proposed overfitting risk level.
    \item Compared to existing approaches, FRAGrow avoids significant accuracy drops in both overfit and underfit cases with a reasonable trade-off in training time.   
\end{itemize}

\section{Related Work}

\subsection{Training acceleration via neural growth}
Most research works that investigate efficient training via neural growth focus on how to initialize new neurons/layers. \cite{Aosong2020Energy-efficientTransformation,Li2022AutomatedTransformers,Dong2020TowardsPerspective}. An early study on new neuron initialization employs the random initialization \cite{Istrate2018IncrementalNetworks}. To accelerate the training of the large neural net, new neurons are initialized via function-preserving transformation to preserve the function and knowledge of the smaller net \cite{Chen2016Net2Net:Transfer}. This function preservation initialization approach is then extended to initialize the new layers or blocks for neural growth in depth \cite{Wei2016NetworkMorphism,Wei2019StableMorphism,Wei2021ModularizedApproach}. To further speed up training, the new weights are initialized to not only preserve the small net's knowledge but also maximize the gradient of new weights \cite{Evci2022GradMax:Information}.  However, research works also report that function initialization may not outperform the random initialization in terms of the final accuracy \cite{Li2022AutomatedTransformers,Wen2020AutoGrow:Networks}. Another line of works initializes the new neurons  by copying the weights from their neighboring neurons \cite{Chang2018Multi-levelView,Dong2020TowardsPerspective} or the historical ensemble of these neighboring weights \cite{Li2022AutomatedTransformers}. This simple method proves effective performance in ResNet \cite{He2016DeepRecognition} and vision transformers \cite{Li2022AutomatedTransformers}, and is therefore utilized in our research.

 Studies \cite{Li2022AutomatedTransformers,Chang2018Multi-levelView} also investigate  ``where to grow" policies. Previous works choose to either interpolate new layers in between the old ones \cite{Chang2018Multi-levelView,Dong2020TowardsPerspective} or stack the new layer after the old ones \cite{Gong2019EfficientStacking}.  A recent approach ``elastic supernet" \cite{Li2022AutomatedTransformers} is built upon the interpolation method, and grows one layer per time by optionally activating new layers and selecting the layer with the highest accuracy.  Nevertheless, these methods either lead to a doubling of the layer count \cite{Chang2018Multi-levelView,Dong2020TowardsPerspective} or need additional computation cost to search the growth location \cite{Li2022AutomatedTransformers} each time the network undergoes expansion. Another line of works simply trains layers sequentially by appending a new layer right to the old layers \cite{Hinton2006ANets,Nkland2019TrainingSignals,Belilovsky2019GreedyImageNet} for higher training efficiency, and this method is employed in our study as the ``where to grow" policy.

Despite these, few studies \cite{Wen2020AutoGrow:Networks} investigate the ``when to grow" policy. Two main policies used in the literature are the periodic growth \cite{Li2022AutomatedTransformers,Chang2018Multi-levelView} and the convergence policy \cite{Istrate2018IncrementalNetworks,Belilovsky2019GreedyImageNet}. Periodic growth employs a fixed growth speed, and convergent growth’s speed is mainly determined by the model’s convergence speed. Without explicitly considering the risks of overfitting and underfitting, these policies might accelerate the training process but could result in a significant reduction in accuracy.

\subsection{Neural architecture search by neural growth}
Neural growth is also used for neural architecture search \cite{Liu2019SplittingArchitectures,Wu2020SteepestSplitting,Wu2020FireflyNetworks,Zhu2020EfficientNetworks,Wen2020AutoGrow:Networks}.  Firefly ~\cite{Wu2020FireflyNetworks} and NeSt ~\cite{Dai2019NeST:Paradigm} grow neurons based on the largest initial gradient. Besides growth location studies, another study  \cite{Wen2020AutoGrow:Networks} also explores growth timing and learning rate scheduling in neural architecture search, and reveals that neural growth shows a preference for fast periodic growth and a constant large learning rate. 

\section{Method}

\subsection{The regularization effect of  neural growth} 

Neural growth exhibits a regularization effect. As Figure \ref{fig:1} (a) shows, the vanilla  method  trains the final large model with layers $B_1$, $B_2$, and $B_3$ directly for $E_1$ epochs. By contrast, neural growth, demonstrated by Figure \ref{fig:1} (b), starts with layer $B_1$ and sequentially inserts and trains layers $B_2$  and $B_3$ for $E_2$ and $E_3$ number of epochs respectively. Due to reduced training epochs ($E_2$ and $E_3$) for the majority of the layers ($B_2$ and $B_3$), their final weight values may be closer to their initial weight values compared to vanilla training, resulting in the regularization effect on the final model. As the regularization effect may increase the training error ~\cite{Goodfellow2016DeepLearning}, one may expect the model trained with neural growth to have higher training error than the model trained by vanilla method without neural growth, and this phenomenon is observed in  Table \ref{method:regularization effect by table}.

\begin{table}[t]
\caption{The regularization effect in neural growth. Compared to slow growth, fast growth employs a smaller growth interval  for the neural network, leading to a more rapid increase in model size and larger average training epochs $\bar{E}$ for fast growth. By contrast, the vanilla method trains the large final model directly without neural growth.  We report the test error (\%) and the training error (\%). }
\label{method:regularization effect by table}
\centering
\begin{tabular}{@{} l l *{2}{c} *{2}{c} @{}}
\toprule
\multirow{2}{*}{Model} & \multirow{2}{*}{Method}
& \multicolumn{2}{c}{CIFAR100} 
& \multicolumn{2}{c@{}}{ImageNet}\\
\cmidrule{3-4} \cmidrule(l){5-6}
& & Test& Train& Test& Train\\
\midrule
\multirow{3}{*}{ResNet}  & Slow growth  &\textbf{28.91} &  6.24  & 24.73 & 21.68 \\
& Fast growth  & 29.33 &  1.82 & 24.29 & 18.48 \\
& Vanilla      &  29.56 &  \textbf{1.12}  & \textbf{24.14} & \textbf{17.82} \\
\midrule
\multirow{3}{*}{VGG}  & Slow growth & 27.34 &  7.40   &  25.70 & 29.47\\
& Fast growth &  \textbf{26.86} &  1.16 & 24.44 & 25.97  \\
& Vanilla      & 26.96 & \textbf{0.32}    & \textbf{24.15} & \textbf{25.05} \\
\bottomrule
\end{tabular}
\end{table}

\begin{figure}[t]
\centering
\includegraphics[width=\linewidth]{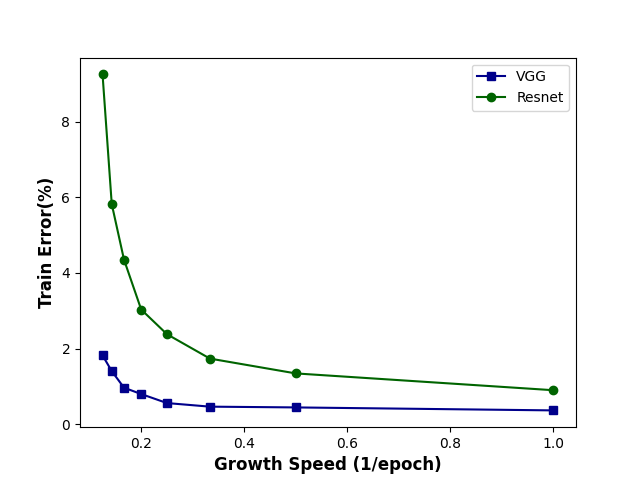}
\vspace{-0.5cm}
\caption{\textbf{Regularization effect of neural growth on CIFAR100 dataset. }  In this experiment, the network is grown periodically, and new network layers are added after each growth interval.  The growth speed is measured by the inverse of the growth interval. Increasing growth speed leads to larger average training epochs $\bar{E}$ and a weaker regularization effect, resulting in smaller training errors.}\label{method: Regularization effect}
\vspace{-0.5cm}
\end{figure}

The  neural growth-induced regularization strength is  determined by the average training epochs $\bar{E}$ in Equation ~\ref{eq:ng tb} where $n$ is the number of added blocks. A smaller $\bar{E}$ may have a stronger regularization effect as shown in Table~\ref{method:regularization effect by table} where the slow growth approach, with larger growth interval and smaller $\bar{E}$, yields higher training errors due to its relatively stronger regularization effect compared to the fast growth approach.  This finding is consistently reaffirmed by Figure ~\ref{method: Regularization effect} where a higher growth speed (larger  $\bar{E}$) consistently leads to decreased training errors, further highlighting the influence of $\bar{E}$.

\begin{equation}
    \bar{E} = \frac{1}{n}\sum_{i=1}^{n}{E_i}\label{eq:ng tb}
\end{equation}

 The impact of the regularization effect from neural growth on the model's generalizability can vary depending on the type of fitting risk present. If the final model overfits the training dataset, the regularization effect from neural growth may improve the model's generalizability. This can be confirmed by the CIFAR100 results in Table ~\ref{method:regularization effect by table} where VGG and ResNet exhibit overfitting on the CIFAR100 dataset, with a low training error (approximately 1\%) but a significantly higher test error (over 25\%). In this context, the regularization effect from neural growth, as shown by slow growth in ResNet and fast growth in VGG, slightly enhances test accuracy. However, when the final target model underfits the training dataset, the regularization effect from neural growth may harm the model's learning capability. For example,  Table ~\ref{method:regularization effect by table} shows both ResNet and VGG models tend to underfit the ImageNet dataset evidenced by the high training error. In this case, the test errors of neural growth approaches (fast and slow growth) are higher than the vanilla method without neural growth. In summary, the choice of an appropriate ``when to grow" policy should be tailored to the  fitting risk type.

\subsection{When to grow policy} 

\begin{table}[t]
\caption{The overfitting risk level (ORL) as the predictor for over/underfitting.  We report the training error (\%) and validation error (\%). }
\label{method:ORL}
\centering
  \setlength{\tabcolsep}{3pt} 
\begin{tabular}{@{} l *{2}{c} *{2}{c} *{2}{c} @{}}
\toprule
\multirow{2}{*}{Model} 
& \multicolumn{2}{c}{CIFAR100} 
& \multicolumn{2}{c@{}}{TinyImageNet}& \multicolumn{2}{c@{}}{ImageNet}\\
\cmidrule{2-3} \cmidrule(l){4-5} \cmidrule(l){6-7}
&  Train  & ORL & Train& ORL &  Train & ORL\\
\midrule
\multirow{1}{*}{ResNet}   &1.12  &31.35 & 0.01  & 45.69&17.82 & 10.76\\
\midrule
\multirow{1}{*}{VGG}   &0.32  & 36.61 & 0.04 & 37.86& 25.05 & 4.30\\
\bottomrule
\end{tabular}
\end{table}

A ``when to grow" policy determines the growth timing ${t_1,...,t_n}$ for n added blocks, which in turn influences the average training epochs $\bar{E}$ and controls the regularization strength. Therefore, it has the potential to mitigate over/under fitting risk of the model by adjusting the growth timing and regularization strength of the model. However, it is necessary to predict the fitting risk type (overfit or underfit) to decide the appropriate growth timing.

To predict overfitting and underfitting, the overfitting risk level (ORL) in equation ~\ref{eq:ORL} is used, which involves comparing the train accuracy and validation accuracy \cite{Bejani2020ConvolutionalSmartphones}. A large ORL suggests that the model may learn excessively irrelevant information from the training dataset, indicating potential overfitting of the model to the dataset. This is evidenced by the CIFAR100 and TinyImageNet ~\cite{Wu2017TinyChallenge} (Table ~\ref{method:ORL}) where the models (ResNet and VGG) overfit these two datasets, with ORL values exceeding 25\%. In contrast, lower ORL values signify good generalization and underfitting risk, demonstrated by ImageNet results in Table ~\ref{method:ORL}, with ResNet and VGG exhibiting around 20\% training error and ORL below 12\%. Thus, ORL during growth is a reliable predictor for identifying the overfitting and underfitting tendencies of the final model.

\begin{equation}
ORL= \text{train accuracy} - \text{validation accuracy} \label{eq:ORL}
\end{equation}

 We propose a simple "when to grow" strategy to mitigate under/over fitting risks based on ORL. It  appends network layers after the dynamic growth interval $I$ specified by Equation ~\ref{eq:when to grow policy} where $I_{max}$ is the maximum allowed growth interval, and $\alpha$ is a hyperparameter.  To allow sufficient time for model finetuning,  $I_{max}$ is constrained by Equation ~\ref{eq:I_max}, where $E_{T}$ is the total train epochs, and $E_{F}^{min}$ is the minimum finetuning epochs, and $n$ is the number of added blocks. When the final model is likely to underfit the dataset via the indication of a small ORL, our ``when to grow" policy Equation ~\ref{eq:when to grow policy}  reduces the growth interval for faster growth, and exerts weaker regularization on the final model. Conversely, if the overfitting risk of the final model on the dataset is detected, the Equation ~\ref{eq:when to grow policy} increases the growth interval, resulting in slower growth and stronger regularization on the final model. Despite its simplicity with only one hyperparameter $\alpha$, we will show its effectiveness in the experiments. A potential tradeoff is that our policy may require extended training time compared to periodic growth with a growth interval of $I_{max}$.

\begin{algorithm}[t]
\caption{Deep Network Growth Algorithm}
\label{alg:neural growth}
\begin{algorithmic}[1]
\STATE \textbf{Input:} A seed shallow network $f_s$; A target deep neural network $f_d$; The number of training epochs $E_T$
\STATE \textbf{Initialization:} The current growing network $f_c=f_s$;
\FOR{$i = 1$ \TO $E_T$} 
    \STATE  Train the model
     \IF{$ |f_c| < |f_d|$ } 
        \IF{When to grow criterion (Eq.~\ref{eq:when to grow policy}) is met} 
             \STATE  Add a new layer at the growth location determined by the ``where to grow" policy 
            \STATE  Initialize the new layer by the Initialization policy
        \ENDIF
    \ELSE
        \STATE finetune the target model $f_d$
    \ENDIF
\ENDFOR
\STATE \textbf{Output:} A trained target deep neural network $f_d$
\end{algorithmic}
\end{algorithm}

\begin{equation}
I=\frac{I_{max}}{{1 + e^{{\alpha-ORL}}}} \label{eq:when to grow policy}
\end{equation}

\begin{equation}
I_{max}=\frac{E_{T}-E^{min}_{F}}{n} \label{eq:I_max}
\end{equation}

\subsubsection{Overhead.} The proposed ``when to grow" policy incurs its main overhead from computing the validation accuracy of ORL in Equation ~\ref{eq:ORL}. However, since the validation accuracy is computed only at the end of every epoch and the validation set size represents only 1\% of the training dataset in our experiments, the overhead is negligible when compared to the overall training time.

\begin{table}[t]
\caption{Effect of learning rate schedule during neural growth. As with slow growth in Table ~\ref{method:regularization effect by table}, we grow the model periodically at random growth locations. Test error (\%) is reported.}
\label{tab:effect-learning-rate-schedule}
\centering
\begin{tabular}{llc}
\toprule
\multirow{1}{*}{Model} & \multirow{1}{*}{Learning rate} & \multicolumn{1}{c}{CIFAR100} \\
\midrule
\multirow{2}{*}{ResNet}
    & Constant & \textbf{28.91} \\
    & Cosine Annealing   & 29.32 \\
\midrule
\multirow{2}{*}{VGG}
    & Constant & \textbf{27.34} \\
    & Cosine Annealing  & 28.06 \\
\bottomrule
\end{tabular}
\end{table}

\section{Experiments}

\subsection{Setup}

\subsubsection{Datasets.} Three datasets, CIFAR10, CIFAR100 ~\cite{Krizhevsky2009LearningImages} and ImageNet ~\cite{Deng2009Imagenet:Database}, are utilized to evaluate the algorithms in this study. The ImageNet dataset encompasses 1,000 classes and comprises a total of 1.28 million training images which are divided into a  training set with 1.27 million images and a validation set with 0.01 million images in our study. The original 50,000 validation images serve as the test set. The images are resized with the smaller edge adjusted to 256 pixels, and subsequently, a crop of size 224x224 is extracted from the resized image. In contrast, CIFAR10 and CIFAR100 are smaller-scale image datasets with 10 and 100 classes respectively, each consisting of 49,500 training images, 500 validation images, and 10,000 test images. The images in CIFAR datasets have a resolution of 32x32 pixels.

\subsubsection{Seed shallow networks and target deep models.} Three representative convolution neural network architectures single-branch network-VGGNet ~\cite{Simonyan2014VeryRecognition}, and multiple-branch networks ResNet ~\cite{He2016DeepRecognition} and MobileNetV2 ~\cite{Sandler2018MobileNetV2:Bottlenecks} are used to evaluate the algorithms. These architectures are composed of multiple stages, wherein each stage stacks several blocks with identical output feature maps. For VGG, the notation VGG-1-1-1-1-1 is used, indicating a VGG model with one VGG block in each of the five stages. Similarly, for ResNet, the notation ResNet-2-2-2-2 is utilized, signifying a ResNet architecture with two residual blocks in each of the four stages. In this study, we investigate the growth of the seed shallow networks  (VGG-1-1-1-1-2, ResNet-2-2-2-2 and MobileNetV2-1-1-1-1-1 architectures) into the target deep models (VGG-2-2-4-4-4, ResNet-8-8-8-8 and MobileNetV2-2-3-4-3-3 architectures). The study utilizes VGG-1-1-1-1-2 to avoid a slow convergence issue when the second block of the final stage is inserted.

\begin{table}[t]
\caption{Training Hyperparameters: Triplet values correspond to ResNet, VGG, and MobileNetV2 (left to right), while the single value remains consistent across all architectures.}
\centering
\begin{tabular}{l|c|c}

 \hline
        & CIFAR10/100  & ImageNet \\

\hline
Initial learning rate &  0.5/0.1/0.1  &  0.1  \\
 \hline
Batch size & 128 & 256 \\
\hline
 Total train epochs $E_T$ &  180 &  120  \\
 \hline
  Weight decay ($10^{-4}$) &  1/5/1 &  1/1/0.4  \\
 \hline
 Optimizer    & \multicolumn{2}{c}{SGD} \\
\hline
    Momentum & \multicolumn{2}{c}{0.9}\\
\hline
Learning rate scheduling &  \multicolumn{2}{c}{cosine decay} \\ 
\hline
Weight initialization & \multicolumn{2}{c}{He \cite{He2015DelvingClassification}} \\
\hline 
Min finetuning epochs $E_F^{min}$ & \multicolumn{2}{c}{30 ~\cite{Dong2020TowardsPerspective}} \\
\hline
\end{tabular}
\label{table1}
\end{table}

\subsubsection{Contenders.} In this study, we utilize three baseline methods: periodic growth, convergent growth and lipgrow ~\cite{Dong2020TowardsPerspective}. As with  the established convention of dividing the overall training procedure into multiple equidistant phases \cite{Li2022AutomatedTransformers,Tan2021EfficientNetV2:Training}, the periodic growth approach in this study involves inserting new layers at a fixed interval of $I_{max}$ in Equation ~\ref{eq:I_max}. The convergent growth method adds blocks only when the accuracy shows stagnation. Lipgrow ~\cite{Dong2020TowardsPerspective} grows the model when the Lipschitz constant of the model exceeds a threshold value. 

\subsubsection{Implementation details.}  For each growth, we append a new block to old blocks in a stage until this stage reaches its target capacity. We then proceed to the later stages until the model attains the target size.  We initialize  the new block by duplicating the weight values from its precedent block \cite{Dong2020TowardsPerspective,Chang2018Multi-levelView}. If the precedent block is a downsampling block, the new block will be randomly initialized ~\cite{Chang2018Multi-levelView,He2015DelvingClassification}. We investigate two learning rate scheduling during growth phase: a constant large learning rate ~\cite{Wen2020AutoGrow:Networks} and cosine annealing with restart ~\cite{Loshchilov2017SGDR:Restarts}. Table ~\ref{tab:effect-learning-rate-schedule} demonstrates that a constant large learning rate outperforms cosine annealing with restart, and thus constant large learning rate is employed ~\cite{Wen2020AutoGrow:Networks} in this study. After the model is grown into the final model, it is finetuned by the cosine decayed learning rate without restart. We employed  the standard data augmentation (horizontal flip and random cropping).  The training hyperparameters are summarized in Table ~\ref{table1}. The experiments are repeated 3 times.

\subsection{Main results}

Table ~\ref{Main Results} compares our neural growth method with vanilla methods that train small  or large  models directly, demonstrating comparable accuracy with the large model baseline while requiring shorter training time. For the CIFAR10/100 dataset, our approach yields approximately 0.3\% improvement in test error for ResNet and VGG, while reducing training time by about 20\% for all models. This accuracy improvement may be attributed to the regularization effect of neural growth on the final target models, VGG-2-2-4-4-4, and ResNet-8-8-8-8, which tend to overfit the CIFAR10/100 dataset. Conversely, on the ImageNet dataset, we observe a slight accuracy drop of around 0.32\%, yet the method saves approximately 10\% of the training time compared to the vanilla approach that directly trains the large target model. Due to MobileNet's 7\% higher training error and increased underfitting on ImagenNet than VGG and ResNet, our policy grows MobileNet faster, resulting in diminished training time savings on MobileNet. Nevertheless, Table ~\ref{when to grow policy} reveals that our accuracy drop is notably smaller than the accuracy reductions observed with the periodic growth (approximately 1.1\%) and convergent growth (approximately 1.7\%) strategies. Finally, it is noteworthy that the time-saving on the CIFAR10/100 dataset surpasses that on the ImageNet dataset due to our method's automatic detection of under/overfitting risks, resulting in slower model growth for the overfitted CIFAR10/100 dataset than for the underfitted ImageNet dataset, and leading to more substantial time savings in the former.

\begin{table}[t]
\caption{Main Results. The terms ``small" and ``large" refer to training a seed shallow network and a target deep model from scratch, respectively, without neural growth. Test error (\%) and normalized training time (\%) are reported.}
\label{Main Results}
\centering
\resizebox{\linewidth}{!}{
\begin{tabular}{@{} l l *{2}{c} *{2}{c} *{2}{c} @{}}
\toprule
\multirow{2}{*}{Model} & \multirow{2}{*}{Method}
& \multicolumn{2}{c}{CIFAR10} 
& \multicolumn{2}{c}{CIFAR100} 
& \multicolumn{2}{c@{}}{ImageNet}\\
\cmidrule{3-4} \cmidrule(l){5-6} \cmidrule(l){7-8}
& & Test & Time& Test& Time & Test& Time\\
\midrule
\multirow{3}{*}{ResNet} & Small & 8.37 & \textbf{39.23} &32.97  &  \textbf{36.95}  & 29.06 & \textbf{58.79} \\
& Large & 6.66 & 100.00 & 29.56 &  100.00  & \textbf{24.14} & 100.00 \\
& Proposed      & \textbf{6.32} & 82.45 &  \textbf{29.14} &  79.93  & 24.32 & 90.00 \\
\midrule
\multirow{3}{*}{VGG}  & Small & 8.27& \textbf{48.21} & 31.12 &   \textbf{51.71} & 31.01 &\textbf{54.35}\\
& Large & 6.22 & 100.00 &  26.96 &   100.00  & \textbf{24.15} & 100.00 \\
& Proposed    & \textbf{6.20} & 81.65 & \textbf{26.57}  & 84.00   & 24.39 & 88.76 \\
\midrule
\multirow{3}{*}{MobileNetV2}  & Small & 7.32 & \textbf{38.12} & 27.49 & \textbf{40.07}   & 36.97 & \textbf{52.70} \\
& Large & \textbf{5.22} & 100.00 & 23.95 &  100.00  & \textbf{29.71} &  100.00\\
& Proposed     & 5.60 & 79.42 & \textbf{23.94} & 73.82   & 30.25 & 93.75 \\
\bottomrule
\end{tabular}
}
\end{table}

\begin{table}[t]
\caption{Comparison of when to grow policies on image classification tasks. Except for the vanilla method that trains the target large model without neural growth, other methods employ neural growth with different when-to-grow policies. Lipgrow ~\cite{Dong2020TowardsPerspective} doubles blocks for each growth. Test error (\%) and normalized training time (\%) are reported. }
\label{when to grow policy}
\centering
  \setlength{\tabcolsep}{4pt}
  \resizebox{\linewidth}{!}{
\begin{tabular}{@{} l l *{2}{c} *{2}{c} *{2}{c} @{}}
\toprule
\multirow{2}{*}{Model} & \multirow{2}{*}{Method}
& \multicolumn{2}{c}{CIFAR10} 
& \multicolumn{2}{c}{CIFAR100} 
& \multicolumn{2}{c@{}}{ImageNet}\\
\cmidrule{3-4} \cmidrule(l){5-6} \cmidrule(l){7-8}
& & Test & Time& Test& Time & Test& Time\\
\midrule
\multirow{5}{*}{ResNet}
& Periodic & 6.58 & 102.33 &  29.29 &  98.62  & 24.79 & 93.39 \\
& Convergent   & 6.35 &  104.39   & 29.25 &  107.42  & 25.30 &  \textbf{86.59}\\
& Lipgrow   & 7.18 &  \textbf{67.22}   & 29.23 &  \textbf{96.09}  & 25.10 & 88.86 \\
& Proposed  & \textbf{6.32} &  100.00    & \textbf{29.14} & 100 .00&  \textbf{24.32} &  100.00 \\
\cmidrule(lr){2-8}
& Vanilla   & 6.66 &  121.29   & 29.56 & 125.11&  24.14 &  111.11 \\
\midrule
\multirow{5}{*}{VGG}
& Periodic & 6.40& 92.23 & 26.73 &  93.02  & 25.70 & 92.40 \\
& Convergent    & 6.33&  99.72  & 26.83 & 92.65   & 26.42 &  \textbf{77.61}\\
& Lipgrow   & 7.05& \textbf{75.92}   & 29.82 &  \textbf{83.26}  & 27.03 &  103.91 \\
& Proposed   & \textbf{6.20} &  100.00    & \textbf{26.57}  & 100.00  & \textbf{24.39} & 100.00  \\
\cmidrule(lr){2-8}
& Vanilla  & 6.22 &   122.48   & 26.96 & 119.05&  24.15 &  112.66 \\
\bottomrule
\end{tabular}
}
\end{table}

\begin{table}[t]
\caption{Comparison of when to grow policies on MobileNetV2~\cite{Sandler2018MobileNetV2:Bottlenecks}. Except for the vanilla method that trains the target large model without neural growth, other methods employ neural growth with different when-to-grow policies. Lipgrow ~\cite{Dong2020TowardsPerspective} doubles blocks for each growth. Test error (\%) and normalized training time (\%) are reported. }
\label{when to grow policy: mobilenet}
\centering
  \setlength{\tabcolsep}{4pt}
  \resizebox{\linewidth}{!}{
\begin{tabular}{@{} l l *{2}{c}  *{2}{c} @{}}
\toprule
\multirow{2}{*}{Model} & \multirow{2}{*}{Method}
& \multicolumn{2}{c}{CIFAR10} 
& \multicolumn{2}{c}{CIFAR100}\\
\cmidrule{3-4} \cmidrule(l){5-6} 
& & Test & Time& Test& Time \\
\midrule
\multirow{5}{*}{MobileNetV2}
& Periodic & 5.66& \textbf{95.11} &  24.35 &  \textbf{99.67}   \\
& Convergent   & \textbf{5.50} &  105.86    &  24.11 &  106.87   \\
& Lipgrow   & 5.64 &  117.30   &  24.32 & 116.65     \\
& Proposed   & 5.60 &  100.00    &  \textbf{23.94} &  100.00    \\
\cmidrule(lr){2-6}
& Vanilla   & 5.22 &  132.92    & 23.95  & 123.83     \\
\bottomrule
\end{tabular}
}
\end{table}

\begin{table}[t]
\caption{Effect of alpha. All neural growth policies are consistent with those described in the method section.  Test error (\%) and normalized training time (\%) are reported. }
\label{Effect of alpha}
\centering
\begin{tabular}{@{} l l *{2}{c} *{2}{c} @{}}
\toprule
\multirow{2}{*}{Model} & \multirow{2}{*}{$\alpha$}
& \multicolumn{2}{c}{CIFAR100} 
& \multicolumn{2}{c@{}}{ImageNet}\\
\cmidrule{3-4} \cmidrule(l){5-6}
& & Test & Time& Test& Time\\
\midrule
\multirow{3}{*}{ResNet}
& 2       & \textbf{29.11} &  99.04  & 24.86 & \textbf{98.23} \\
& 4 (Ours) & 29.14 &  100.00  & 24.32 & 100.00 \\
& 6       & 29.20 &   \textbf{97.70} & \textbf{24.27} & 100.90 \\
\midrule
\multirow{3}{*}{VGG}
& 2       & 26.75 &  \textbf{95.11}  & \textbf{24.22} &  101.20 \\
& 4 (Ours) & \textbf{26.57} &  100.00  & 24.39 & \textbf{100.00} \\
& 6       & 26.89 &  110.58  &  24.32 & 99.17 \\
\bottomrule
\end{tabular}
\end{table}

\begin{table}[t]
\caption{Effect of where to grow policies. The ``when to grow," ``how much to grow," and initialization policies remain consistent with those outlined in the method section. Test error (\%) and normalized training time (\%) are reported. }
\label{effect of where to grow policy}
\centering
\begin{tabular}{@{} l l *{2}{c} *{2}{c} @{}}
\toprule
\multirow{2}{*}{Model} & \multirow{2}{*}{Method}
& \multicolumn{2}{c}{CIFAR100} 
& \multicolumn{2}{c@{}}{ImageNet}\\
\cmidrule{3-4} \cmidrule(l){5-6}
& & Test & Time& Test& Time\\
\midrule
\multirow{3}{*}{ResNet}
& Periodic & 28.41 & 101.30  & 24.62 & \textbf{98.20}  \\
& Convergent       & 29.15 &  107.91  & 25.07 & 99.13 \\
& Proposed       & \textbf{28.39} &  \textbf{100.00}  &  \textbf{24.52} &  100.00\\
\midrule
\multirow{3}{*}{VGG}
& Periodic & 27.00 &  \textbf{85.74}  & 26.19 &  94.85 \\
& Convergent       & 27.00 & 95.10  & 26.32 &  \textbf{94.78} \\
& Proposed        &  \textbf{26.63} &  100.00 & \textbf{24.18} & 100.00 \\
\bottomrule
\end{tabular}
\end{table}

Table ~\ref{when to grow policy} and Table ~\ref{when to grow policy: mobilenet} show the proposed method  achieves superior accuracy compared to existing approaches in the case of underfitting while exhibiting similar accuracies in overfitting scenarios. On the overfitting CIFAR10/100 dataset, all methods except lipgrow~\cite{Dong2020TowardsPerspective} achieve comparable accuracy with the target large model, indicating their appropriate growth speed and regularization strengths through neural growth.  On the underfitting ImageNet dataset, our method outperforms the strongest baseline method, periodic growth, by 0.47\% and 1.31\% for ResNet and VGG architectures, respectively. This improvement results from our method's ability to detect underfitting risks and increase growth speed accordingly, reducing the regularization impact of neural growth. In contrast, while periodic growth and convergent growth exhibit suitable growth speeds for the overfitting CIFAR10/100 dataset, their growth speeds prove too slow for underfitting ImageNet dataset, imposing excessive regularization on the model and leading to lower accuracies. The fast growth speed is particularly necessary for the VGG model on the ImageNet dataset due to the more pronounced underfitting risk compared to ResNet. Nevertheless, it should be noted that our method requires longer training time than periodic growth, as the growth interval in our method never exceeds that of periodic growth, resulting in faster growth speed and less training time saved compared to periodic growth.

\subsection{Ablation study}

We study two aspects of our method, aiming to answer the following questions: How does the choice
of hyperparameters impact the performance of
FRAGrow? How robust is our growth timing policy to growth order and initialization of new layers of neural growth?

\subsubsection{Effect of hyperparameter $\alpha$.} We investigate the impact of  $\alpha$  in equation ~\ref{eq:when to grow policy} that determines the growth interval dynamically. Results in Table ~\ref{Effect of alpha} show that reducing $\alpha$ from 6 to 2 led to fluctuating test errors or a drop in accuracy (around 0.6\% for ResNet on ImageNet). The accuracy decline is attributed to the slower growth speed with smaller $\alpha$, resulting in a stronger regularization effect and reduced accuracy on underfitting ImageNet datasets. However, the slower growth with a smaller $\alpha$ saves more training time, as confirmed by VGG on CIFAR100. Therefore, an optimal $\alpha$ value should balance training efficiency and regularization strength, with our experiments indicating that an $\alpha$ value of 4 performed well for the considered datasets.

\subsubsection{Effect of growth order} To assess the influence of the ``where to grow" policy,  we replace the layerwise growth with the circulation growth ~\cite{Wen2020AutoGrow:Networks} that iterates over the stages and append a new block to a stage whenever it is visited. The experimental results are presented in Table ~\ref{effect of where to grow policy}. The proposed approach exhibits robustness to the growth order and achieves comparable or higher accuracy compared to baseline methods (periodic growth and convergent growth). Particularly noteworthy is the experiment involving the VGG model on the underfitting ImageNet dataset, where the proposed method shows an accuracy improvement of approximately 2\% over its contenders. This outcome reinforces the significance of increasing growth speed for underfitting models.

\subsubsection{Effect of initialization} To evaluate the impact of new layers initialization, we employ the moment growth ~\cite{Li2022AutomatedTransformers}, involving copying the weights of new layers from the historical ensemble of its preceding layer. The results are presented in Table ~\ref{effect of initialization}. Our method demonstrates comparable accuracy to baseline techniques on the overfitting CIFAR100 dataset while exhibiting approximately 1\% higher average accuracy on the underfitting ImageNet dataset. This underscores the resilience of our method towards various initialization strategies. It should be noted that our proposed approach exhibits a marginal improvement in time efficiency compared to the baseline methods on the ImageNet dataset. This disparity arises due to the intrinsic requirement of the moment growth technique to calculate moving averages of historical weight values throughout each training step during model expansion, leading to additional computational overhead.  As the proposed method grows the model fast on the ImageNet dataset, the associated extra computational burden becomes comparatively diminished, yielding a slightly reduced training time compared to the baseline methods. 

\begin{table}
\caption{Effect of initialization. 
The ``when to grow," ``where to grow," and ``how much to grow" policies remain in line with the method section's description. Test error (\%) and normalized training time (\%) are reported. }
\label{effect of initialization}
\centering
\begin{tabular}{@{} l l *{2}{c} *{2}{c} @{}}
\toprule
\multirow{2}{*}{Model} & \multirow{2}{*}{Method}
& \multicolumn{2}{c}{CIFAR100} 
& \multicolumn{2}{c@{}}{ImageNet}\\
\cmidrule{3-4} \cmidrule(l){5-6}
& & Test & Time& Test& Time\\
\midrule
\multirow{3}{*}{ResNet}
& Periodic & 28.94 & \textbf{97.47}  & 24.72 & 105.01 \\
& Convergent       & 29.82 & 105.80 & 25.28   & 101.12 \\
& Proposed       & \textbf{28.80} &  100.00  & \textbf{24.65} & \textbf{100.00} \\
\midrule
\multirow{3}{*}{VGG}
& Periodic & 26.85 &  93.43  & 26.15 & 106.74 \\
& Convergent       & 26.86 & \textbf{89.27}  & 26.52 & 105.01 \\
& Proposed        &  \textbf{26.67} & 100.00  & \textbf{24.24} & \textbf{100.00} \\
\bottomrule
\end{tabular}
\end{table}

\section{Conclusion}
In this paper, we have investigated the growth timing policy to address under/overfitting risks. Specifically, we have discovered that neural growth induces a regularization effect on the target large model, with the regularization strength being governed by the ``when to grow" policy. To address these risks effectively, we introduced the concept of overfitting risk level, allowing us to predict and manage the under/overfitting risks for the target large model by adjusting the growth speed accordingly. Through experimental evaluations on image recognition tasks involving both single-branch networks (VGG) and multiple-branch networks (ResNet), we demonstrated the efficacy of our proposed FRAGrow method. As part of future work,  we intend to extend our method to more vision tasks, e.g., dense prediction tasks, and improve the training efficiency of these tasks.

\section{Acknowledgements} 
This study was supported by the Melbourne Graduate Research Scholarship for the first author and the Australian Research Council grant DP220101035 for the first, third, fourth, and last authors. The research utilized the LIEF HPC-GPGPU Facility at the University of Melbourne, established with support from LIEF Grant LE170100200.

\bibliography{main}


\end{document}